\documentclass[a4paper,fleqn]{cas-sc}

\usepackage[numbers]{natbib}
\usepackage[section]{placeins}

\def\tsc#1{\csdef{#1}{\textsc{\lowercase{#1}}\xspace}}
\tsc{WGM}
\tsc{QE}
\tsc{EP}
\tsc{PMS}
\tsc{BEC}
\tsc{DE}

\makeatletter
\ExplSyntaxOn
\RenewDocumentCommand{\process@marks}{}
   {
     \cs_if_free:cTF { mark@corau\theauthor }
       { \ignorespaces }
       { \str_set:Nx \l_tmpa_str { \use:c{ mark@corau\theauthor } }
         \int_case:nn { \l_tmpa_str }
           {
             { 1 } { \sep$\dagger$ }
             { 2 } { \sep$\ddagger$ }
             { 3 } { \sep$\dagger\!\dagger$ }
           }
         \tex_def:D \sep{\unskip,}
       }
     \cs_if_free:cTF { mark@fnau\theauthor }
       { \ignorespaces }
       { \str_set:Nx \l_tmpa_str { \use:c{ mark@fnau\theauthor } }
         \int_case:nn { \l_tmpa_str }
           {
             { 1 } { \sep$\ast$ }
             { 2 } { \sep$\ast\!\ast$ }
             { 3 } { \sep$\ast\!\ast\!\ast$ }
           }
         \tex_def:D \sep{\unskip,}
       }
   }

\RenewDocumentCommand \fntext { O{} m }
{
  \tl_if_head_eq_catcode:nNTF { #1 } a
  {
    \seq_gput_right:Nn \g_stm_fnote_seq
    { \int_incr:N \g_stm_fnote_int
      \str_set:Nx \@currentlabel { \int_use:N \g_stm_fnote_int }
      \stmLabel { #1 }
      \tex_def:D \thefootnote
      { \int_case:nn { \g_stm_fnote_int }
        {
          { 1 } { $\ast$ }
          { 2 } { $\ast\!\ast$ }
          { 3 } { $\ast\!\ast\!\ast$ }
        }
      }
      \footnotetext { #2 }
    }
  }
  {
    \seq_gput_right:Nn \g_stm_fnote_seq
    {
      \int_set:Nn \l_tmpa_int { #1 }
      \tex_def:D \thefootnote
      { \int_case:nn { \l_tmpa_int }
        {
          { 1 } { $\ast$ }
          { 2 } { $\ast\!\ast$ }
          { 3 } { $\ast\!\ast\!\ast$ }
        }
      }
      \footnotetext { #2 }
    }
  }
}

\RenewDocumentCommand \cortext { O{} m }
{
  \tl_if_head_eq_catcode:nNTF { #1 } a
  {
    \seq_gput_right:Nn \g_stm_cor_seq
    { \int_incr:N \g_stm_cor_int
      \str_set:Nx \@currentlabel { \int_use:N \g_stm_cor_int }
      \stmLabel { #1 }
      \tex_def:D \thefootnote
      { \int_case:nn { \g_stm_cor_int }
        {
          { 1 } { $\dagger$ }
          { 2 } { $\ddagger$ }
          { 3 } { $\dagger\!\dagger$ }
        }
      }
      \footnotetext { #2 }
    }
  }
  {
    \seq_gput_right:Nn \g_stm_cor_seq
    {
      \int_set:Nn \l_tmpa_int { #1 }
      \tex_def:D \thefootnote
      { \int_case:nn { \l_tmpa_int }
        {
          { 1 } { $\dagger$ }
          { 2 } { $\ddagger$ }
          { 3 } { $\dagger\!\dagger$ }
        }
      }
      \footnotetext { #2 }
    }
  }
}
\ExplSyntaxOff
\makeatother

\begin{document}
\let\WriteBookmarks\relax
\def\floatpagepagefraction{1}
\def\textpagefraction{.001}
\shorttitle{Echo-DM: Ultrasound Marker Removal via Conditional Latent Diffusion and Region-Aware Fusion}
\shortauthors{Wang et~al.}

\title [mode = title]{Echo-DM: Ultrasound Marker Removal via Conditional Latent Diffusion and Region-Aware Fusion}                      


\author[1]{Zhiwei Wang}
\ead{zhiweiwanghh@whu.edu.cn}
\fnref{fn1}

\author[1]{Tao Huang}
\ead{taohuang@whu.edu.cn}
\fnref{fn1}

\author[1]{Wentao Jiang}

\author[2]{Muyi Li}

\author[2]{Jianxin Liu}

\author[2]{Jian Chen}
\corref{cor1}

\author[3]{Jie Zou}

\author[1]{Yong Luo}

\author[1]{Bo Du}

\author[1]{Jing Zhang}
\ead{jingzhang.cv@whu.edu.cn}
\corref{cor1}

\address[1]{School of Computer Science, Wuhan University, China}
\address[2]{The Central Hospital of Wuhan, China}
\address[3]{School of Computer Science, Hubei University of Technology, China}

\cortext[cor1]{Corresponding author}
\fntext[fn1]{These authors contributed equally to this work.}

\begin{abstract}
Clinical ultrasound images often contain artificial markers, such as measurement calipers and text, to assist diagnostic interpretation and comparison. However, these markers can introduce shortcut bias in downstream automated analysis, encouraging deep learning models to rely on marker-related cues rather than clinically meaningful anatomy. Existing marker removal methods are either mask-dependent and vulnerable to error propagation, or mask-free deterministic restorers that may over-smooth ultrasound texture and perturb unaffected background regions. To address these challenges, we present \textbf{Echo-DM}, a framework for ultrasound marker removal via conditional latent diffusion and region-aware fusion. Echo-DM follows a common encoder-diffusion-decoder pipeline, where a DiT-based conditional latent diffusion network performs global restoration and a region-aware fusion module enforces preservation-aware image-space refinement under end-to-end mask-free inference. Building on this fixed core design, we further instantiate \textbf{Echo-DM-V} and \textbf{Echo-DM-R} with VAE-based and RAE-based latent modules, respectively, which demonstrates that the Echo-DM architecture is compatible with diverse latent-module instantiations. Extensive experiments on \textbf{Echo-PAIR}, a large-scale paired clinical ultrasound dataset, demonstrate superior marker removal and strong anatomical fidelity compared with representative two-stage baselines, while providing favorable quality-efficiency trade-offs across deployment settings. Data, code and models will be released at \href{https://github.com/MiliLab/Echo-DM}{Echo-DM}.
\end{abstract}




\maketitle
\section{Introduction}
\begin{figure}[pos=htbp]
    \centering
    \includegraphics[width=0.95\linewidth]{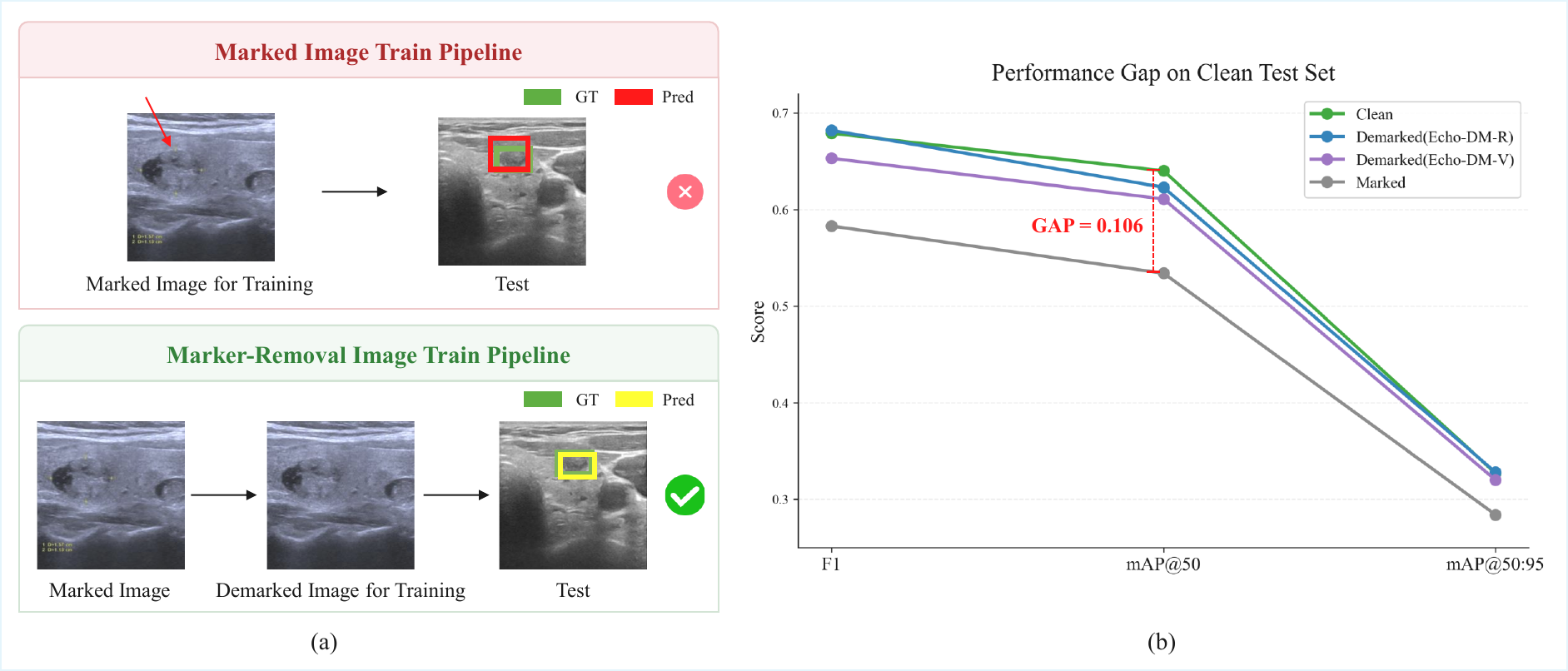}
    \caption{Motivating evidence for marker-induced train--deployment mismatch in ultrasound analysis. (a) Qualitative example on a shared clean test image: models trained on marked images show localization drift, whereas using de-marked training images improves detection quality. (b) Quantitative comparison of different training-data constructions evaluated on the same clean test set. Models trained on images processed by Echo-DM-V or Echo-DM-R substantially reduce the gap to clean-image training compared with models trained directly on marked images.}
    \label{fig:motivation_shortcut_bias}
\end{figure}

In routine clinical practice, ultrasound images often contain overlaid markers, such as measurement calipers, crosshairs, and textual notes, added to localize lesions and record size-related information. As a result, marked images account for a substantial proportion of hospital historical ultrasound archives, rather than native clean images. Although these annotations are useful for follow-up examination and diagnostic comparison, directly using such marked data to train downstream models can easily introduce shortcut bias into practical automated workflows \citep{YaoEtAl2020TextureSynthesis}. Instead of learning clinically meaningful pathological patterns from the underlying tissue, neural networks may rely excessively on these high-contrast marker cues. As illustrated in Fig.~\ref{fig:motivation_shortcut_bias}, training on marked ultrasound images can degrade downstream performance on clean inputs, whereas de-marker preprocessing helps narrow this gap. Therefore, developing an accurate and high-fidelity marker removal method for ultrasound image preprocessing is of substantial importance for building reliable and generalizable downstream medical AI systems \citep{YaoEtAl2020TextureSynthesis,YingEtAl2020CMR,LiEtAl2024MMI}.

Existing methods for ultrasound marker removal can be broadly grouped into three categories: cascaded mask-based restoration, end-to-end deterministic reconstruction, and more recent diffusion-based restoration \citep{YingEtAl2020CMR,LiEtAl2024MMI,GuoEtAl2023BlindInpainting,ZhangEtAl2024UltrasonicNoise2Noise}. Mask-based pipelines provide explicit regional control by first localizing marker regions and then restoring them, but their performance depends heavily on mask quality and is vulnerable to error propagation \citep{YingEtAl2020CMR,LiEtAl2024MMI}. End-to-end deterministic methods avoid external masks and simplify inference, yet they often struggle to preserve fine ultrasound texture and unaffected background under deterministic reconstruction objectives \citep{GuoEtAl2023BlindInpainting,ZhangEtAl2024UltrasonicNoise2Noise,LedigEtAl2017SRGAN}. More recent diffusion-based approaches offer a stronger generative prior, but faithful marker removal remains difficult when intact anatomical regions must be preserved as much as possible \citep{LugmayrEtAl2022RePaint,RombachEtAl2022LDM,HansenEtAl2024LumbarMRIInpainting,ProchazkaZeman2026StableDiffusionUS}. As a result, current methods still struggle to remove markers without introducing unnecessary changes to intact anatomical content.

To address this dilemma and achieve \emph{minimal necessary editing}, we propose \textbf{Echo-DM}, a framework for ultrasound marker removal via conditional latent diffusion and region-aware fusion. Echo-DM follows a unified encoder-diffusion-decoder (EDD) pipeline, where a Diffusion Transformer (DiT) \citep{PeeblesXie2023DiT} performs conditional latent restoration under the guidance of the marked-image latent representation, and a Region-Aware Fusion (RAF) Module further refines the restored image in pixel space to preserve intact background regions. Under the same restoration-and-fusion core, Echo-DM supports different latent encoder-decoder instantiations and is compatible with different latent-representation choices, e.g., VAE-based and RAE-based latent modules \citep{KingmaWelling2014VAE,ZhengEtAl2025RAE}. This unified design enables restoration of marker-corrupted inputs toward a clean anatomical appearance while limiting unnecessary modifications to unaffected regions, and supports end-to-end mask-free inference at deployment.

To support a systematic evaluation of our method, we establish \textbf{Echo-PAIR}, a large-scale paired clinical ultrasound dataset for marker removal. Extensive experiments show that Echo-DM delivers strong performance in both local restoration quality and global fidelity. Our main contributions are summarized below:
\begin{itemize}
    \item We identify a marker-induced \emph{domain gap} between marked ultrasound images and clean anatomical appearance, and propose an Echo-DM model family to reduce this gap so that downstream models are less affected by overlaid markers.
    
    \item We show that the Echo-DM framework is compatible with both VAE-based and RAE-based latent encoder-decoder modules, enabling practical quality--efficiency trade-offs without changing the main restoration pipeline.
    
    \item We establish \textbf{Echo-PAIR}, a large-scale and diverse paired clinical ultrasound dataset, and demonstrate strong restoration performance of Echo-DM in both marker-region recovery and full-image anatomical fidelity.
\end{itemize}

\section{Related Work}
Existing studies on ultrasound marker removal can be broadly grouped into three categories: two-stage mask-based restoration, end-to-end deterministic restoration, and diffusion-based restoration \citep{YingEtAl2020CMR,LiEtAl2024MMI,GuoEtAl2023BlindInpainting,ZhangEtAl2024UltrasonicNoise2Noise}. These categories differ mainly in how corrupted regions are localized and how restoration is balanced against background fidelity \citep{YingEtAl2020CMR,LiEtAl2024MMI,GuoEtAl2023BlindInpainting,RombachEtAl2022LDM}.

\subsection{Two-Stage Mask-Based Restoration}
Early work on ultrasound marker removal mainly followed a two-stage pipeline that first localizes the corrupted region and then restores the masked content. In the broader image restoration literature, partial convolution and gated convolution established representative mask-conditioned inpainting formulations for irregular holes, providing a practical technical basis for subsequent small-region restoration pipelines \citep{LiuEtAl2018PConv,YuEtAl2019GatedConv}. In the ultrasound domain, Ying et al. presented a cascaded marker-removal method that explicitly decomposed the task into marker extraction and subsequent erasure for thyroid ultrasound images \citep{YingEtAl2020CMR}. Sun et al. used an edge-connection algorithm to detect manually induced artifacts, followed by Criminisi-based image restoration on the detected regions \citep{SunEtAl2021EdgeCriminisi}. More recently, Li et al. combined learned binary marker-mask prediction with mask-guided inpainting to build a more streamlined and task-specific detection--erasure pipeline for small and salient artificial markers \citep{LiEtAl2024MMI}.

These methods provide explicit regional control, which is attractive for small and structured markers. Their restoration quality, however, remains strongly dependent on the accuracy of explicit masks, which motivates later work on mask-free restoration pipelines.

\subsection{End-to-End Deterministic Restoration}
To reduce reliance on explicit masks, later studies explored end-to-end deterministic restoration for direct marker removal from marked images. One representative line reformulates the task as mask-free blind inpainting. Building on generic blind inpainting formulations \citep{WangEtAl2020VCNet}, Guo et al. proposed a two-branch reconstruction network that implicitly localizes corrupted regions while restoring the image, and further improves local fidelity with object-aware discrimination \citep{GuoEtAl2023BlindInpainting}. In a related blind visible watermark removal setting, Meng et al. proposed a dual-pathway framework that jointly performs spatial positioning and background restoration under end-to-end mask-free inference, and showed that feature fusion can improve both localization and restoration quality \citep{meng2025dfcl}. Another line treats annotations as a noise source and performs self-supervised restoration without requiring clean targets. A recent ultrasound Noise2Noise-style study follows this idea and learns annotation removal directly from noisy-only supervision \citep{LehtinenEtAl2018Noise2Noise,ZhangEtAl2024UltrasonicNoise2Noise}.

Compared with two-stage pipelines, these methods are more convenient in practice because they support end-to-end mask-free inference. Once localization is learned only implicitly, preserving intact background regions becomes more difficult; moreover, deterministic reconstruction objectives may over-smooth fine-grained ultrasound texture, thereby reducing background fidelity \citep{LedigEtAl2017SRGAN}.

\subsection{Diffusion-Based Restoration}
Diffusion-based restoration introduces a stronger generative prior for recovering corrupted content and has become increasingly relevant to medical image editing and inpainting. In the generic image domain, Lugmayr et al. and Kawar et al. demonstrated the potential of diffusion-based restoration for inpainting and broader inverse problems \citep{LugmayrEtAl2022RePaint,KawarEtAl2022DDRM}. In medical imaging, related studies have applied diffusion models to pathology editing, image enhancement and denoising, and ultrasound inpainting. Hansen et al. highlighted the importance of integrating generated content with surrounding anatomy in lumbar spine MRI inpainting \citep{HansenEtAl2024LumbarMRIInpainting}. Ben Alaya et al. further explored counterfactual diffusion-based image editing on brain MRI \citep{BenAlayaEtAl2025MedEdit}, while Yuan et al. extended conditional latent diffusion to medical image enhancement \citep{YuanEtAl2025MedIENet}. Gong et al. also showed that diffusion models can support denoising-oriented medical restoration beyond pure editing or synthesis \citep{GongEtAl2024PETDDPM}. More broadly, recent medical-vision studies suggest that diffusion models can support not only image generation or editing, but also noise-robust representation learning and structured prediction: DiffCNN combines a diffusion subnet with a CNN subnet for semi-supervised medical image segmentation \citep{xu2025diffcnn}, while recent Stable-Diffusion-based ultrasound work shows that latent diffusion has started to enter ultrasound inpainting applications \citep{ProchazkaZeman2026StableDiffusionUS}.

Beyond these application-level differences, the latent encoder-decoder module is also an important design factor in latent diffusion pipelines. Most existing systems are built on VAE-style latent interfaces \citep{KingmaWelling2014VAE,RombachEtAl2022LDM}, while recent work has explored Representation Autoencoders (RAEs) in diffusion transformer settings \citep{ZhengEtAl2025RAE}. These studies suggest that different latent representation modules can lead to different quality--efficiency operating characteristics. Related encoder-decoder studies further indicate that simple skip connections may be insufficient to reconcile semantic gaps across multi-scale features, motivating more learnable feature interaction beyond direct feature copying \citep{wang2024narrowing}. 

However, diffusion-based studies on faithful ultrasound marker removal remain limited, and existing medical diffusion work is still more often centered on editing, synthetic augmentation, inpainting, enhancement, or denoising than on faithful recovery of marker-occluded anatomy. This leaves room for methods that combine the stronger generative prior of diffusion restoration with stricter control over minimal necessary editing and background preservation.

\section{Methods}
\subsection{Problem Formulation and Overview}
\label{sec:method-overview}
Given an ultrasound image with artificial markers, denoted by $x_m$, our goal is to recover the corresponding clean image $\hat{x}$ while preserving the texture and structural integrity of the unaffected anatomical background. Unlike conventional cascaded methods that rely on explicit marker masks, we aim to develop a high-fidelity marker removal method that requires no external mask during inference. In addition to effectively removing local occlusions, the model should avoid introducing unnecessary modifications to background regions that are meant to remain unchanged.

\begin{figure}[pos=htbp]
    \centering
    \includegraphics[width=\linewidth]{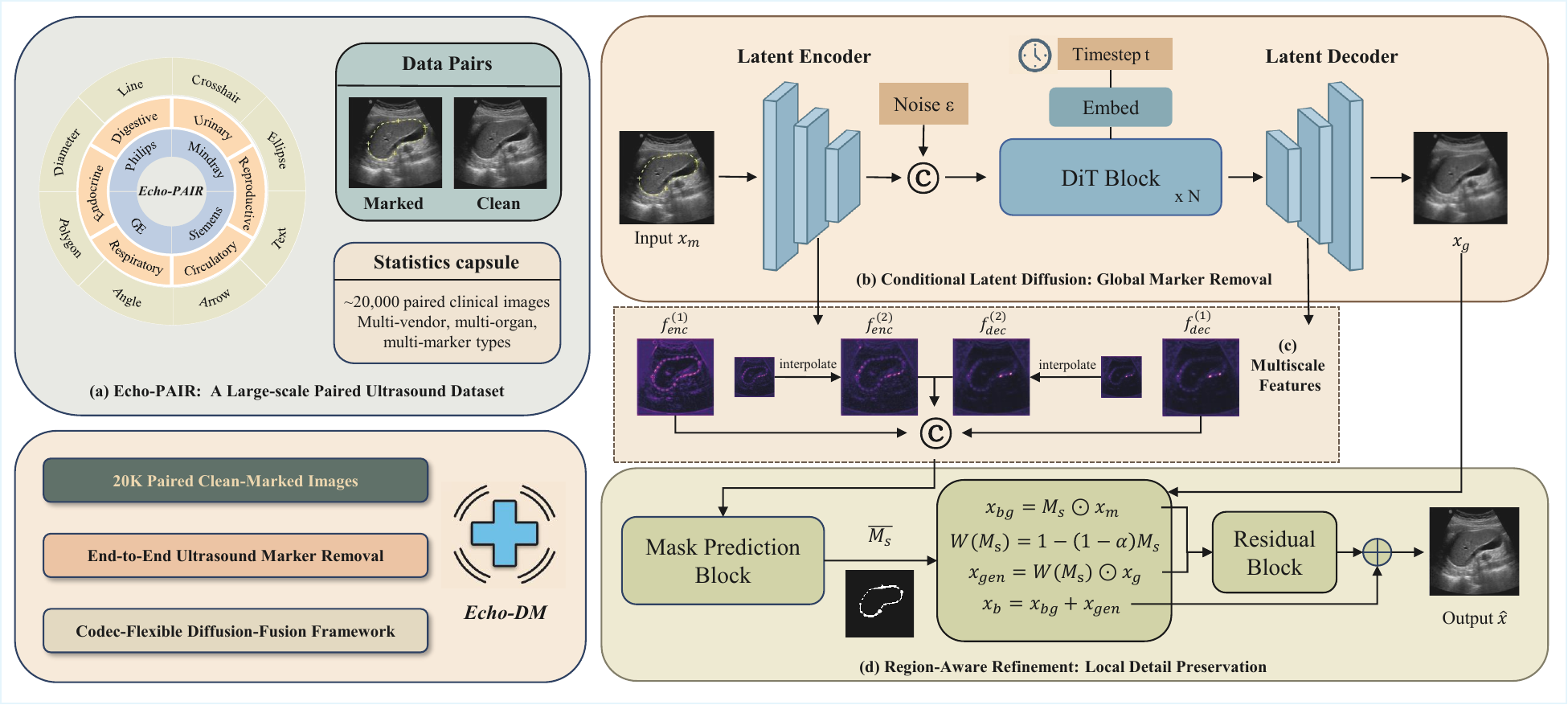}
    \caption{Overall framework of Echo-DM. (a) Echo-PAIR: A Large-scale Paired Ultrasound Dataset. Echo-PAIR provides about 20K paired clean-marked ultrasound images covering multi-vendor, multi-organ, and multi-marker scenarios, supporting end-to-end mask-free inference in a codec-flexible diffusion-fusion framework. (b) Conditional Latent Diffusion: Global Marker Removal. The marked input $x_m$ is encoded into latent space; with timestep conditioning and Gaussian noise, a DiT denoiser performs latent restoration and the decoder produces a globally restored image $x_g$. (c) Multiscale Features. Shallow encoder and late decoder features (e.g., $f_{\mathrm{enc}}^{(1)}$, $f_{\mathrm{enc}}^{(2)}$, $f_{\mathrm{dec}}^{(2)}$, $f_{\mathrm{dec}}^{(1)}$) are aligned by interpolation and aggregated to predict the fusion mask, where their discrepancies are most pronounced around marker-affected regions. (d) Region-Aware Refinement: Local Detail Preservation. A mask-prediction branch estimates a soft map $M_s$ to adaptively fuse preserved background from $x_m$ with generated content from $x_g$, and a residual block further corrects local details to produce the final output $\hat{x}$.}
    \label{fig:echo_dm_main_method}
\end{figure}

As shown in Fig.~\ref{fig:echo_dm_main_method}, \textbf{Echo-DM} is built on an ultrasound-adapted latent encoder-decoder module and includes two task-specific components: a shared Conditional Latent Diffusion Network (CLDN) and a RAF Module. The adapted latent module provides latent encoding and decoding that are better matched to ultrasound textures and anatomical structures. On this basis, the CLDN performs global restoration in latent space within a latent-diffusion framework \citep{RombachEtAl2022LDM}, using the marked image as the direct visual condition. The RAF Module then refines the restored result in image space through region-aware fidelity fusion, aiming to preserve the intact background from the input while improving local consistency and boundary naturalness in the final output.

Overall, Echo-DM does not treat ultrasound marker removal as a generic image inpainting problem. Instead, it is designed to jointly satisfy two complementary objectives: sufficient marker removal and necessary background preservation. The former is achieved by conditional latent restoration, while the latter is further strengthened by image-space region-aware fusion. During training, Echo-DM is optimized in a stage-wise manner, where the latent encoder-decoder module is first adapted to the ultrasound domain to establish reliable latent encoding and decoding, followed by the learning of latent restoration and region-aware fusion. During inference, the entire method operates as an end-to-end marker removal pipeline without requiring any explicit external mask. Although soft-mask supervision is used during training to guide the fusion branch (Section~\ref{sec:method-stagewise}), no external mask is provided at inference time. Section~\ref{sec:method-vae} introduces ultrasound-domain latent module adaptation, Section~\ref{sec:method-dit} details the CLDN, Section~\ref{sec:method-fusion} presents the RAF Module, and Section~\ref{sec:method-stagewise} describes the stage-wise optimization and end-to-end mask-free inference procedure.

\subsection{Ultrasound-domain Latent Module Adaptation}
\label{sec:method-vae}
Echo-DM operates in latent space. Therefore, the quality of the latent representation and its decoding process directly determines the upper bound of downstream marker removal. Following latent-diffusion practice \citep{RombachEtAl2022LDM}, we use a latent encoder-decoder module to map images between pixel space and latent space. This module can be instantiated with either VAE-based or RAE-based designs \citep{KingmaWelling2014VAE,ZhengEtAl2025RAE}. However, in ultrasound imaging, generic pretrained latent modules are often suboptimal for domain-specific texture characteristics and anatomical structures, especially high-frequency speckle patterns that are critical for faithful decoding.

To establish more suitable latent encoding and decoding for ultrasound marker removal, we adapt a pretrained latent encoder-decoder module to the ultrasound domain before training the restoration network. This step is not an additional task branch; instead, it serves as task-specific representation alignment for the subsequent conditional diffusion and image-space fusion stages. Similar domain-adaptation strategies for latent diffusion pipelines have also been explored in medical imaging \citep{ProchazkaZeman2026StableDiffusionUS}.

\paragraph{Instantiation-specific initialization.}
For the VAE branch, we initialize from a natural-image pretrained VAE and then adapt it on the target ultrasound training set. For the RAE branch, we initialize from MAE-based pretrained weights \citep{HeEtAl2022MAE}, and then continue adaptation on the same target ultrasound training set used by the VAE branch. This design keeps the downstream diffusion and fusion training stages comparable while isolating differences in latent-module instantiation and initialization.

Specifically, in our default adaptation protocol we retain the encoder and fine-tune only the decoder. This design preserves latent-space stability for conditional diffusion learning while improving reconstruction fidelity in ultrasound appearance space. It also reduces trainable parameters and mitigates overfitting risk compared with fully updating both encoder and decoder. The detailed optimization objective of this stage is presented in Section~\ref{sec:method-stagewise}. In the remainder of the method, the resulting ultrasound-adapted latent module is used to encode both marked and clean images into latent representations and to decode restored latents back to image space.

\subsection{Conditional Latent Diffusion Network}
\label{sec:method-dit}
To avoid reducing ultrasound marker removal to deterministic pixel-wise regression, we adopt a Diffusion Transformer (DiT) as the denoising network of a latent-space conditional diffusion model \citep{PeeblesXie2023DiT}, and tailor it to the image-to-image restoration setting. Given a marked ultrasound image $x_m$ and its corresponding clean image $x_c$, we first map them into the ultrasound-adapted latent space using the encoder of the adapted latent module:
\begin{equation}
z_m = \mathcal{E}(x_m), \qquad z_c = \mathcal{E}(x_c),
\end{equation}
where $z_m$ serves as the conditional latent, and $z_c$ denotes the clean-image latent representation to be restored.

Rather than treating marker removal as unconditional generation or mask-guided inpainting, our goal is to model a conditional latent restoration process that recovers a clean anatomical latent under the guidance of the marked-image latent. To this end, we retain the standard Transformer architecture of DiT while adapting its conditioning mechanism to this task. Specifically, the original class-conditioning branch is removed, and only the timestep embedding is preserved as the diffusion modulation signal. Meanwhile, instead of introducing the marked-image latent $z_m$ through a separate condition encoder or a cross-attention branch, we directly concatenate it with the current latent state $z_t$ along the channel dimension \citep{SahariaEtAl2021SR3} to form a joint input:
\begin{equation}
\tilde{z}_t = [z_t; z_m].
\end{equation}
The joint latent $\tilde{z}_t$ is then tokenized and passed to the DiT denoising network for noise prediction.

Based on this design, the conditional diffusion denoising network is formulated as
\begin{equation}
\hat{\epsilon}_{\theta} = f_{\theta}(z_t, z_m, t),
\end{equation}
where $f_{\theta}(\cdot)$ denotes the parameterized DiT denoising network, $t$ is the diffusion timestep, and $\hat{\epsilon}_{\theta}$ is the predicted noise. Here, $z_t$ denotes the current latent state along the diffusion trajectory, while $z_m$ provides the direct visual condition from the marked image. The diffusion forward process and optimization objective are detailed in Section~\ref{sec:method-stagewise}.

A key advantage of this formulation is that it does not rely on any explicit external marker mask. Instead, it directly exploits the visible tissue information and contextual structure preserved in the latent space of the marked image to condition the restoration of occluded regions. Compared with deterministic full-image reconstruction optimized only with pixel-wise objectives, latent diffusion modeling can better preserve plausible high-frequency and stochastic texture characteristics in ultrasound images, while alleviating over-smoothing tendencies.

Overall, the role of the CLDN is to use the latent representation of the marked image as a direct visual condition and restore a global result in latent space that is closer to the clean anatomical distribution. This component provides a semantically informative and structurally consistent restoration foundation for the subsequent RAF Module, which further enhances background fidelity and boundary naturalness in image space.

\subsection{Region-Aware Fusion Module}
\label{sec:method-fusion}
Although the CLDN can recover a global result closer to the clean-image distribution, latent-space decoding alone may still introduce subtle texture drift or boundary inconsistencies in unaffected regions \citep{RombachEtAl2022LDM,ProchazkaZeman2026StableDiffusionUS}. This issue is particularly important in ultrasound imaging, where high-frequency speckle patterns carry meaningful anatomical information. To mitigate it, we introduce a RAF Module that performs fidelity-oriented fusion in image space, aiming to preserve the background content of the input image that should remain unchanged.

Specifically, let $x_g$ denote the globally restored image produced by the conditional diffusion denoising network after decoding. Rather than regenerating the entire image uniformly, the fusion module follows the principle of \emph{minimal necessary editing} by combining the preservable background from the input image with the content restored by the generative branch. To this end, we aggregate encoder/decoder features across available scales, denoted by
\begin{equation}
F = \{f_{\mathrm{enc}}^{(k)},\, f_{\mathrm{dec}}^{(k)}\}_{k=1}^{K}.
\end{equation}
Here, the number of feature levels $K$ depends on the latent module instantiation. For the VAE-based instantiation (Echo-DM-V), we use two shallow encoder levels and two late decoder levels (equivalently, \(K=2\): \(f_{\mathrm{enc}}^{(1)}, f_{\mathrm{enc}}^{(2)}, f_{\mathrm{dec}}^{(1)}, f_{\mathrm{dec}}^{(2)}\)). For the RAE-based instantiation (Echo-DM-R), explicit multi-scale hierarchy is not available in the same form, so we use a single feature level (\(K=1\)). Based on these features, a gating network estimates a soft fusion mask:
\begin{equation}
M_s = \mathcal{G}(F),
\end{equation}
where $M_s \in [0,1]^{H \times W}$ indicates the degree of background preservation at each spatial location. Unlike the explicit segmentation masks used in conventional two-stage methods \citep{YingEtAl2020CMR,LiEtAl2024MMI}, $M_s$ is not intended to delineate marker boundaries independently. Instead, it serves as an internal guidance signal that adaptively controls the fusion ratio between the original background and the generated content.

Using this soft fusion mask, we first construct the background-preserving branch:
\begin{equation}
x_{\mathrm{bg}} = M_s \odot x_m,
\end{equation}
and the generative restoration branch:
\begin{equation}
x_{\mathrm{gen}} = W(M_s) \odot x_g,
\end{equation}
where
\begin{equation}
W(M_s) = 1 - (1-\alpha) M_s.
\end{equation}
Here, $\alpha \in [0,1]$ controls the minimum contribution of the generative branch in regions with high background-retention confidence. In regions where $M_s$ takes low values, the model relies more heavily on the generated content to complete the restoration. Conversely, in regions where $M_s$ takes high values, the original background is preferentially preserved, while a small contribution from the generative branch is still retained to alleviate discontinuities caused by hard switching near boundaries. The fidelity-preserving base image is then obtained as
\begin{equation}
x_b = x_{\mathrm{bg}} + x_{\mathrm{gen}}.
\end{equation}

To further reduce unnatural local transitions or residual fusion artifacts, we introduce a lightweight residual refinement branch that performs a small corrective adjustment on the base image, yielding the final output:
\begin{equation}
\hat{x} = x_b + R(x_b).
\end{equation}
To better satisfy the design goal of \emph{minimal necessary editing}, the output layer of the residual branch is zero-initialized, so that the network is biased at the beginning of training toward preserving the physically interpretable structure of the base image and learns local corrections only when necessary \citep{ZhangEtAl2023ControlNet}.

Overall, the RAF Module combines the global restoration capability of the CLDN with the high-fidelity background information contained in the input image itself, thereby further improving background consistency and boundary naturalness in the final result. Compared with directly decoding the entire image in a uniform manner, this module is better suited to ultrasound marker removal, as it explicitly emphasizes the preservation of unaffected regions and restricts model edits to areas that truly require restoration.

\subsection{Stage-wise Optimization and End-to-End Mask-Free Inference}
\label{sec:method-stagewise}
Since Echo-DM consists of two task-specific components serving different purposes, namely latent-space conditional restoration and image-space region-aware fusion, we adopt a stage-wise optimization strategy for training. During inference, the entire method performs marker removal in an end-to-end manner without any explicit external mask. This design allows each component to converge stably under the supervision signal that is most suitable for its role, while maintaining a unified input-output pipeline for practical deployment.

\paragraph{Stage I: Ultrasound-domain latent module adaptation.}
We first adapt the latent encoder-decoder module to the ultrasound domain in order to obtain a latent representation and decoding capability that are better suited to this task. This stage is trained on ultrasound images without requiring clean--marked pairing at the sample level, while model selection is performed on a paired validation subset (clean/marked) to evaluate reconstruction behavior under both appearances. Starting from a pretrained latent module, our default protocol freezes the encoder and fine-tunes only the decoder, which preserves latent-space stability for downstream conditional diffusion learning.

The training objective is defined as
\begin{equation}
\mathcal{L}_{\mathrm{latent}}
=
\lambda_{1}\mathcal{L}_{1}
+
\lambda_{p}\mathcal{L}_{\mathrm{LPIPS}}
+
\lambda_{\mathrm{reg}}\mathcal{L}_{\mathrm{reg}},
\end{equation}
where $\mathcal{L}_{1}$ enforces pixel-wise consistency between the decoded image and the input image, $\mathcal{L}_{\mathrm{LPIPS}}$ enhances perceptual texture fidelity, and $\mathcal{L}_{\mathrm{reg}}$ denotes the latent regularization term of the chosen encoder-decoder instantiation (e.g., KL regularization for VAE-based instantiation) \citep{ZhangEtAl2018LPIPS,KingmaWelling2014VAE,ZhengEtAl2025RAE}. During validation, we monitor decoding quality on both clean and marked inputs, and select the checkpoint according to validation SSIM on marked reconstructions to better reflect the target deployment condition. Overall, this stage performs domain adaptation of latent encoding and decoding rather than direct marker-removal learning, and provides a stabilized representation basis for Stage II.

\paragraph{Stage II: Conditional diffusion optimization.}
We then train the CLDN. Given paired marked and clean images, both are first mapped into the adapted latent space, where the marked latent $z_m$ serves as the conditional input and the clean latent $z_c$ serves as the diffusion target. Valid-region masks are used during training to exclude padded areas from attention and supervision. For a randomly sampled timestep $t$, the forward diffusion process is defined as \citep{HoEtAl2020DDPM}
\begin{equation}
z_t
=
\sqrt{\bar{\alpha}_t}\, z_c
+
\sqrt{1-\bar{\alpha}_t}\,\epsilon,
\qquad
\epsilon \sim \mathcal{N}(0, I),
\end{equation}
and the corresponding training objective of the conditional diffusion denoising network is
\begin{equation}
\mathcal{L}_{\mathrm{DiT}}
=
\mathbb{E}_{z_c, z_m, \epsilon, t}
\left[
\left\|
\epsilon - \hat{\epsilon}_{\theta}(z_t, z_m, t)
\right\|_2^2
\right].
\end{equation}
This stage is responsible for learning a conditional restoration mapping from marked observations to the clean anatomical distribution, thereby establishing global structural restoration capability. For model selection, we choose checkpoints according to validation performance.

\paragraph{Stage III: Region-aware fusion fine-tuning.}
Finally, we further fine-tune the RAF Module using paired samples \((x_m, x_c)\), where \(x_m\) is the marked input participating in fusion and \(x_c\) provides clean-image supervision and evaluation reference. At this point, the underlying adapted latent representation module remains frozen, and optimization is applied only to fusion-related parameters (skip adapter, feature gate, fusion block, and decoder output head).

To reduce the error propagation from imperfect DiT predictions during this stage, we construct the fusion condition latent from the clean image as
\begin{equation}
z_{\mathrm{cond}} = \mathcal{E}(x_c) + \sigma \epsilon, \qquad \epsilon \sim \mathcal{N}(0, I),
\end{equation}
instead of using the DiT-predicted latent. This decouples fusion learning from diffusion prediction errors and allows Stage II and Stage III training to be organized with lower inter-stage dependency.

The overall objective is defined as
\begin{equation}
\mathcal{L}_{\mathrm{fusion}}
=
\lambda_{r}\mathcal{L}_{\mathrm{rec}}
+
\lambda_{p}\mathcal{L}_{\mathrm{LPIPS}}
+
\lambda_{m}\mathcal{L}_{\mathrm{mask}},
\end{equation}
where \(\mathcal{L}_{\mathrm{rec}}\) denotes a region-weighted reconstruction loss:
\begin{equation}
\mathcal{L}_{\mathrm{rec}}
=
\left\|
\left(1+\beta M_{\mathrm{mark}}^{d}\right)\odot (\hat{x}-x_c)
\right\|_1,
\end{equation}
with \(M_{\mathrm{mark}}^{d}\) denoting a dilated marker-region mask derived from \(|x_c-x_m|\). This weighting imposes stronger penalties on marker-related regions while keeping global consistency constraints. \(\mathcal{L}_{\mathrm{LPIPS}}\) preserves perceptual texture consistency \citep{ZhangEtAl2018LPIPS}. \(\mathcal{L}_{\mathrm{mask}}\) supervises the gating branch to learn a reasonable allocation between background preservation and region replacement. The soft-mask supervision term is further defined as
\begin{equation}
\mathcal{L}_{\mathrm{mask}}
=
\mathcal{L}_{\mathrm{BCE}}
+
\mathcal{L}_{\mathrm{Dice}}.
\end{equation}
In practice, \(\mathcal{L}_{\mathrm{BCE}}\) is applied to background-retention prediction (target \(1-M_{\mathrm{mark}}^{d}\)), while \(\mathcal{L}_{\mathrm{Dice}}\) is applied to its complementary marker-region prediction, improving both pixel-level and region-level mask learning. The key hyperparameters are defined accordingly: \(\alpha\) controls the minimum contribution of the generative branch in high-confidence background regions (to avoid hard-switch artifacts), \(\sigma\) controls the perturbation scale of \(z_{\mathrm{cond}}\), the dilation radius controls boundary coverage of \(M_{\mathrm{mark}}^{d}\), and \(\lambda_r,\lambda_p,\lambda_m\) balance marker removal strength and background fidelity.

\paragraph{Inference: End-to-end mask-free deployment.}
During inference, given a marked ultrasound image, the model first restores the corresponding global latent result through the CLDN, after which the RAF Module combines the original background information from the input image with the output of the restoration branch to produce the final result. No explicit external mask is required throughout the inference pipeline, allowing the method to perform high-fidelity marker removal directly on real ultrasound images.

In addition, inference can optionally use a high-resolution fusion strategy. Under this setting, the conditional diffusion denoising network performs global restoration at a standard resolution, while the fusion module produces the final output at a higher resolution. This design keeps the inference cost largely manageable and helps reduce information loss caused by image resizing, further improving the fidelity of local details and background textures.

\section{Results}
\subsection{Dataset and Evaluation Protocol}
\label{sec:dataset-eval-protocol}

\paragraph{Dataset.}
We conduct experiments on \textbf{Echo-PAIR}, a large-scale paired clinical ultrasound dataset for marker removal. The dataset contains approximately \textbf{20{,}000} marked--clean image pairs (about \textbf{40{,}000} images in total), covering multiple ultrasound vendors, anatomical targets, and marker types. The markers mainly include electronic crosshair calipers, dotted linear measurement guides, dotted circular measurement contours, and textual dimension annotations used for lesion assessment. Data were collected from several mainstream clinical ultrasound systems, including Mindray DC-80/DC-80S, GE Voluson E10/E8, Philips EPIQ 7/EPIQ 7C, and Siemens Sequoia Silver. In addition, Echo-PAIR covers representative clinical categories, including abdominal organs (e.g., liver, gallbladder, pancreas, spleen, kidneys, and bladder), obstetric scans (early pregnancy and mid/late-pregnancy fetal views), gynecologic scans (e.g., uterus), superficial-organ scans, and male urologic scans (e.g., prostate and testis), providing a diverse and clinically representative benchmark for ultrasound marker removal.

\paragraph{Evaluation protocol.}
Following practical deployment requirements, we evaluate both restoration quality and inference efficiency. Echo-PAIR is randomly partitioned into training, validation, and test subsets. The validation subset contains \textbf{100 marked--clean image pairs}, and the held-out test subset contains \textbf{300 marked--clean image pairs}; all remaining pairs are used for training. Model selection is performed on the validation subset, while the test subset is reserved strictly for final reporting. For quantitative comparison, we report full-image fidelity and marker-region fidelity simultaneously, with quality metrics computed at the original image resolution. In addition, we report average inference time (seconds per image) to assess whether the method remains within an acceptable runtime range for clinical preprocessing scenarios.

\paragraph{Quality metrics.}
We report PSNR and SSIM on the full image, together with ROI-restricted variants (PSNR-ROI and SSIM-ROI) on marker-affected regions. This combination jointly characterizes global photometric fidelity and local restoration quality in clinically relevant regions. Let $x\in[0,255]^{H\times W\times 3}$ be the clean image and $\hat{x}\in[0,255]^{H\times W\times 3}$ be the restored image.
\begin{equation}
\mathrm{MSE}(x,\hat{x})=\frac{1}{3HW}\sum_{c=1}^{3}\sum_{i=1}^{H}\sum_{j=1}^{W}\left(x_{ijc}-\hat{x}_{ijc}\right)^2,
\end{equation}
\begin{equation}
\mathrm{PSNR}(x,\hat{x})=10\log_{10}\left(\frac{255^2}{\mathrm{MSE}(x,\hat{x})}\right).
\end{equation}
PSNR is an error-sensitivity metric derived from pixel-wise distortion energy, and is widely used for objective fidelity comparison \citep{HuynhThuGhanbari2008PSNR,WangBovik2009MSE}. However, as a pure error-based measure, it may not fully reflect perceived structural quality across heterogeneous image content \citep{WangBovik2009MSE}. For SSIM, we follow the standard definition \citep{WangEtAl2004SSIM}, which emphasizes structural consistency:
\begin{equation}
\mathrm{SSIM}(x,\hat{x})=\frac{1}{HW}\sum_{i=1}^{H}\sum_{j=1}^{W} S_{ij},
\end{equation}
\begin{equation}
S_{ij}=
\frac{(2\mu_x\mu_{\hat{x}}+C_1)(2\sigma_{x\hat{x}}+C_2)}
{(\mu_x^2+\mu_{\hat{x}}^2+C_1)(\sigma_x^2+\sigma_{\hat{x}}^2+C_2)},
\end{equation}
where \(\mu_x,\mu_{\hat{x}}\) are local means, \(\sigma_x^2,\sigma_{\hat{x}}^2\) are local variances, and \(\sigma_{x\hat{x}}\) is local covariance. In practice, we compute SSIM on RGB images with \(data\_range=255\) and report the image-level mean. Therefore, PSNR and SSIM provide complementary views: the former captures pixel-level numerical fidelity, while the latter better reflects local structural consistency.

\paragraph{ROI definition.}
To evaluate marker-region restoration, we define two complementary masks for each test image: (i) a GT Mask derived from paired marked/clean image differencing in Echo-PAIR to indicate true marker-affected pixels, and (ii) a Pred Mask obtained from an nnU-Net model trained on Echo-PAIR marker segmentation annotations. Both masks are aligned to the target image resolution and binarized. The final ROI mask is defined as the union of the two masks followed by one-step \(3\times3\) dilation:
\begin{equation}
\Omega = \mathrm{Dilate}\!\left(M_{\mathrm{GT}} \cup M_{\mathrm{Pred}}\right).
\end{equation}
This design improves robustness to imperfect mask boundaries and potential under-coverage of either source. If one mask source is unavailable, the available mask is used.

\paragraph{ROI metrics.}
ROI-PSNR is computed from RGB MSE within \(\Omega\) after normalizing pixel differences to \([0,1]\):
\begin{equation}
\mathrm{MSE}_{\Omega}=\frac{1}{3|\Omega|}\sum_{c=1}^{3}\sum_{(i,j)\in\Omega}\left(\frac{\hat{x}_{ijc}-x_{ijc}}{255}\right)^2,
\end{equation}
\begin{equation}
\mathrm{PSNR\text{-}ROI}=-10\log_{10}\!\left(\mathrm{MSE}_{\Omega}+10^{-8}\right).
\end{equation}
For ROI-SSIM, we first obtain the full-image SSIM map \(S\), then report the masked average over \(\Omega\):
\begin{equation}
\mathrm{SSIM\text{-}ROI}=\frac{1}{|\Omega|}\sum_{(i,j)\in\Omega}S_{ij},
\end{equation}
with channel averaging applied when \(S\) is represented per-channel.

\paragraph{Efficiency metric.}
We report end-to-end inference latency as seconds per image:
\begin{equation}
T_{\mathrm{img}}=\frac{1}{N}\sum_{n=1}^{N}t_n,
\end{equation}
where $t_n$ is the wall-clock runtime for sample $n$ and includes the full Echo-DM inference path (latent encoding, diffusion sampling, fusion, and decoding). Lower values indicate better efficiency.

\subsection{Implementation Details}

All Echo-DM variants follow the stage-wise optimization and mask-free inference protocol described in Section~\ref{sec:method-stagewise}, with shared latent-module adaptation, conditional diffusion optimization, and region-aware fusion fine-tuning. We therefore report only the implementation details necessary for reproducibility in the main text. Unless otherwise specified, all models are implemented in PyTorch 2.7.1 and optimized using AdamW, and checkpoint selection is performed on the predefined Echo-PAIR validation split (100 image pairs) in Section~\ref{sec:dataset-eval-protocol}. In this paper, Echo-DM-V denotes the VAE-based instantiation, while Echo-DM-R denotes the RAE-based instantiation; both share the same DiT restoration and fusion framework unless otherwise specified.

Configuration-dependent settings, including GPU allocation, input resolution, per-variant hyperparameters, and deployment-oriented variant configurations, are summarized in the appendix for completeness.

\subsection{Main Quantitative Comparison}

Following the evaluation protocol in Section~\ref{sec:dataset-eval-protocol}, all methods are evaluated on the held-out \textbf{Echo-PAIR} test set, with quality metrics computed at the original image resolution.

Table~\ref{tab:main_quantitative_comparison} presents the main quantitative comparison against two representative two-stage mask-guided diffusion inpainting baselines, together with two mask-free baselines that serve as the foundations of our two Echo-DM variants. In both two-stage baselines, a pretrained nnU-Net \citep{IsenseeEtAl2021nnUNet} is first used to predict marker masks from marked ultrasound images, and the resulting mask is then used by a Stable Diffusion inpainting model (v1.5 or v2, both built on latent diffusion \citep{RombachEtAl2022LDM}) to restore the masked region. To characterize the impact of mask source on two-stage pipelines, each baseline is evaluated under three inference-mask settings: Pred mask, GT mask, and Pred $\cup$ GT mask. For a direct comparison under the fixed Echo-DM framework, we further report two mask-free baselines: a DiT baseline without the RAF Module, which serves as the direct baseline of Echo-DM-V, and a plain RAE baseline, which serves as the corresponding baseline of Echo-DM-R. Unless otherwise stated, the default Echo-DM-V setting uses a training resolution of \(512\times512\), a latent compression factor of 4, 50 diffusion sampling steps, and high-resolution fusion at inference.

As shown in Table~\ref{tab:main_quantitative_comparison}, the proposed Echo-DM achieves the strongest restoration performance while retaining a fully mask-free inference protocol. Compared with the two-stage Stable Diffusion inpainting baselines, both Echo-DM variants deliver consistently better restoration quality without relying on an explicit test-time mask. In contrast, the two-stage baselines remain sensitive to the choice of inference mask, with stronger ROI restoration under GT-mask guidance than under predicted masks, indicating that localization errors from the first stage are propagated to the subsequent inpainting stage and constrain the final restoration quality. Within the Echo-DM family, Echo-DM-V is more favorable for full-image fidelity and structural consistency, whereas Echo-DM-R provides a more efficient alternative with stronger local error correction.

Compared with their respective mask-free baselines, Echo-DM-V and Echo-DM-R show clear improvements in background retention while also taking marker removal into account, indicating that the gain comes not only from the latent module itself but from the proposed Echo-DM design as a whole. The conditional latent diffusion backbone restores anatomically plausible content in marker-corrupted regions, while the RAF Module suppresses unnecessary modifications in intact areas. As a result, the Echo-DM series concentrates edits around truly affected areas while maintaining high background fidelity.

\begin{table}[pos=htbp]
\caption{Main quantitative comparison on the Echo-PAIR test set. `Pred mask' denotes the marker mask predicted by the nnU-Net model trained on Echo-PAIR annotations; `GT mask' denotes the marker mask obtained from paired marked/clean differencing; and `Pred $\cup$ GT mask' denotes their union. DiT and all Echo-DM variants are mask-free at inference. For each inpainting backbone, runtime is independent of mask source and is therefore reported once per model.}
\label{tab:main_quantitative_comparison}

\centering
\footnotesize
\setlength{\tabcolsep}{4pt}
\renewcommand{\arraystretch}{1.12}
\begin{tabular*}{\tblwidth}{@{\extracolsep{\fill}} l l c c c c c @{}}
\toprule
Method & Inference setting & Time (s/image)$\downarrow$ & PSNR$\uparrow$ & SSIM$\uparrow$ & PSNR-ROI$\uparrow$ & SSIM-ROI$\uparrow$ \\
\midrule
\multirow{3}{*}{SD v1.5 Inpainting \citep{RombachEtAl2022LDM}}
& Pred mask             & \multirow{3}{*}{3.327} & 29.6862 & 0.8961 & 20.6822 & 0.7092 \\
& GT mask               &                        & 29.8622 & 0.8964 & 22.0487 & 0.7479 \\
& Pred $\cup$ GT mask   &                        & 29.7249 & 0.8962 & 20.9634 & 0.7141 \\
\midrule
\multirow{3}{*}{SD v2 Inpainting \citep{RombachEtAl2022LDM}}
& Pred mask             & \multirow{3}{*}{2.896} & 30.4055 & 0.8677 & 20.6609 & 0.6250 \\
& GT mask               &                        & 30.6287 & 0.8674 & 22.0563 & 0.6663 \\
& Pred $\cup$ GT mask   &                        & 30.4689 & 0.8673 & 21.0261 & 0.6337 \\
\midrule
DiT \citep{PeeblesXie2023DiT}        & Mask-free & 1.168 & 26.075 & 0.8934 & 26.6971 & 0.7983 \\
RAE \citep{ZhengEtAl2025RAE}        & Mask-free & \textbf{0.560} & 23.7236 & 0.7681 & 22.0511 &  0.7225 \\
\midrule
\textbf{Echo-DM-V (Ours)} & Mask-free & 1.237 & \textbf{40.5753} & \textbf{0.9861} & 26.6898 & \textbf{0.8277} \\
\textbf{Echo-DM-R (Ours)} & Mask-free & 0.618 & 31.7483 & 0.9499 & \textbf{27.683} & 0.8218 \\
\bottomrule
\end{tabular*}
\end{table}

\subsection{Analysis}
\label{sec:results-analysis}

\subsubsection{Qualitative Comparison of Marker-Region Reconstruction}
\label{sec:qualitative-analysis}

\begin{figure}[pos=htbp]
\centering
\includegraphics[width=\linewidth]{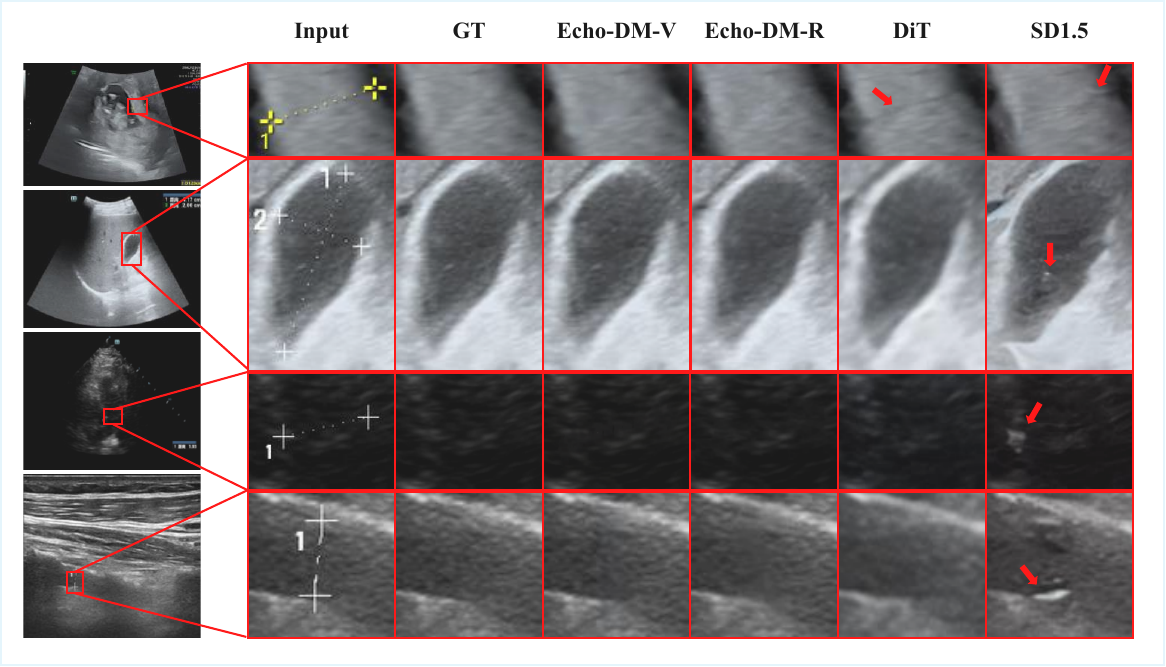}
\caption{Qualitative comparison of marker-region reconstruction across methods. Four representative marker-affected regions from Echo-PAIR are shown, with columns corresponding to the marked input, clean GT, Echo-DM-V, Echo-DM-R, the DiT baseline, and SD v1.5 inpainting. Echo-DM-V and Echo-DM-R remove markers while better preserving local ultrasound texture and structural continuity. By comparison, the DiT baseline shows blurrier appearances around restored regions, and SD v1.5 inpainting exhibits both texture blur and occasional restoration artifacts.}
\label{fig:qualitative_analysis}
\end{figure}

To complement the quantitative comparison, we provide qualitative analysis in Fig.~\ref{fig:qualitative_analysis} using representative test cases from Echo-PAIR. The figure focuses on marker-affected regions and compares the marked input, the clean GT, the DiT baseline, SD v1.5 inpainting, and the two Echo-DM variants under the same display setting.

Across all four examples, Echo-DM-V and Echo-DM-R remove the overlaid markers while maintaining local texture patterns that remain visually closer to the clean GT. Their reconstructed regions show better speckle continuity and more natural structural transitions around the previously corrupted areas, indicating that the proposed restoration-and-fusion design can recover marker-occluded content without introducing obvious local inconsistencies.

By comparison, the DiT baseline, which lacks the RAF Module, tends to produce blurrier appearances around the restored regions, suggesting weaker preservation of local image details after marker removal. SD v1.5 inpainting further exhibits both texture blur and occasional restoration artifacts, leading to less faithful ultrasound appearance than the Echo-DM variants.

\subsubsection{Multiscale Feature Analysis and Soft-Mask Rationality}
\label{sec:mask-analysis}

\begin{figure}[pos=htbp]
\centering
\includegraphics[width=\linewidth]{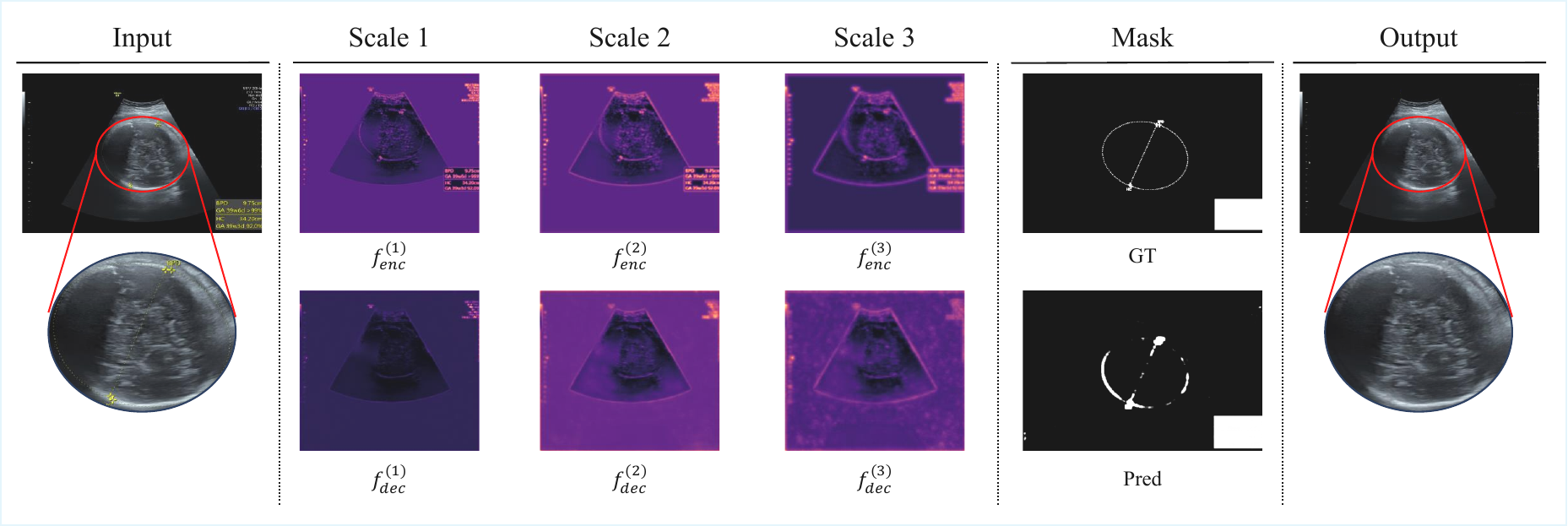}
\caption{Multiscale feature, soft-mask, and output analysis in Echo-DM-V. \textbf{Feature stage}: encoder and decoder features at selected levels are interpolated to a unified size and concatenated as the input to mask prediction; encoder responses emphasize marker-corrupted regions, while decoder responses shift after restoration, and their discrepancy provides implicit localization cues. \textbf{Mask stage}: the predicted soft mask is broadly consistent with the GT mask in major marker-region coverage (white denotes marker regions), while mild boundary deviations remain. \textbf{Output stage}: when coupled with region-aware residual fusion, the resulting output removes marker artifacts while preserving locally coherent texture and boundary continuity.}
\label{fig:mask_analysis}
\end{figure}

To further analyze why region-aware fusion is effective, Fig.~\ref{fig:mask_analysis} presents multiscale encoder/decoder features, the predicted soft mask, the GT mask (white denotes marker regions in the masks), and the final restored output. This analysis is conducted on Echo-DM-V, where explicit encoder-decoder multiscale hierarchy is available. Features from different levels are first interpolated to a unified spatial size and then concatenated before being fed into the mask-prediction module.

From the feature-response patterns, encoder features still exhibit strong activations around marker-corrupted regions, reflecting the presence of overlaid artifacts in the input representation. After conditional diffusion restoration and decoding, decoder-side responses shift because these marker regions are reconstructed toward anatomically plausible content. The response discrepancy between encoder and decoder features therefore provides implicit localization cues for where replacement is needed during fusion.

We also observe that very small-scale features become noticeably blurred after interpolation, which weakens boundary sharpness and can reduce mask-prediction precision. For this reason, the fusion module uses only the first two relatively high-resolution levels for multiscale aggregation, trading off contextual coverage and spatial fidelity.

Finally, the predicted soft mask is largely consistent with the GT mask in marker-region coverage, while remaining smoother at boundaries. The final restored output further shows that this mask-guided fusion design, together with the residual refinement branch, can suppress marker artifacts while maintaining locally coherent texture and boundary continuity.

\subsection{Ablation Study on the RAF Module}
\label{sec:ablation_fusion}

To further validate the effectiveness of the proposed RAF Module and the stability of key design choices, we conduct grouped ablations on \textbf{Echo-DM-V}. Unless otherwise specified, all variants are trained at \(512\times512\), with a VAE compression factor of 4 and 50 diffusion sampling steps at inference. The default Echo-DM-V setting uses noise perturbation strength \(\sigma_n=0.05\), fusion weight \(\alpha=0.1\), and high-resolution fusion at inference. Here, \(\sigma_n\) denotes the standard deviation of Gaussian perturbation injected into the latent input used for fusion training. In each ablation, only one factor is changed while all others are kept unchanged. To avoid test-set leakage during model design, all ablation results in this subsection are reported on the predefined Echo-PAIR validation split (100 marked--clean pairs).

\begin{table}[pos=htbp]
\caption{Case-based ablation results on Echo-DM-V, evaluated on the Echo-PAIR validation split (100 marked--clean pairs). The default setting uses high-resolution fusion at inference, noise perturbation strength \(\sigma_n=0.05\), and fusion weight \(\alpha=0.1\).}
\label{tab:ablation_fusion}
\centering
\small
\setlength{\tabcolsep}{4pt}
\renewcommand{\arraystretch}{1.12}
\begin{tabular*}{\tblwidth}{@{\extracolsep{\fill}} l c c c c c c c @{}}
\toprule
\multirow{2}{*}{Case} & \multicolumn{3}{c}{Configuration} & \multicolumn{4}{c@{}}{Validation metrics} \\
\cmidrule(lr){2-4}\cmidrule(l){5-8}
& \textbf{HR} & \textbf{\(\sigma_n\)} & \textbf{\(\alpha\)} & PSNR$\uparrow$ & SSIM$\uparrow$ & PSNR-ROI$\uparrow$ & SSIM-ROI$\uparrow$ \\
\midrule
Default & \checkmark & 0.05 & 0.1 & \textbf{41.5439} & 0.9845 & 27.6628 & \textbf{0.8307} \\
(a) &  & 0.05 & 0.1 & 29.5010 & 0.9587 & 27.0437 & 0.8255 \\
(b) & \checkmark & 0.03 & 0.1 & 39.8126 & 0.9847 & 27.1207 & 0.8044 \\
(c) & \checkmark & 0.08 & 0.1 & 41.1159 & \textbf{0.9910} & \textbf{27.7318} & 0.8226 \\
(d) & \checkmark & 0.05 & 0 & 40.6441 & 0.9816 & 27.2559 & 0.8281 \\
(e) & \checkmark & 0.05 & 0.2 & 37.8711 & 0.9712 & 27.2959 & 0.7921 \\
\bottomrule
\end{tabular*}
\end{table}

As shown in Table~\ref{tab:ablation_fusion}, the validation-set ablations are consistent with the design rationale of Echo-DM-V. Disabling high-resolution inference causes a marked drop in full-image fidelity (PSNR/SSIM), and also lowers ROI metrics. This behavior suggests that high-resolution inference is important not only for global texture fidelity, but also for stabilizing local restoration quality around marker-affected regions.

For the noise perturbation strength (\(\sigma_n\)), the results indicate a non-monotonic trade-off. Increasing \(\sigma_n\) to 0.08 yields the highest SSIM and PSNR-ROI, but its SSIM-ROI remains below the default setting. In contrast, \(\sigma_n=0.03\) underperforms on both PSNR and ROI metrics. Therefore, \(\sigma_n=0.05\) remains a balanced operating point, providing the highest full-image PSNR together with the best ROI-SSIM.

For the fusion weight, \(\alpha\) directly controls the relative influence of the generative branch in the final fusion. An overly large \(\alpha\) (\(0.2\)) leads to clear degradation, especially on SSIM and SSIM-ROI, suggesting over-injection of generated content into regions that should be preserved. Compared with \(\alpha=0\), the default \(\alpha=0.1\) gives better results across all reported metrics, indicating that a small but non-zero generative contribution is beneficial for preserving continuity while maintaining background fidelity.

\subsection{Quality--Efficiency Trade-off of Echo-DM Variants}
\label{sec:quality_efficiency}

Beyond the main comparison and module-level ablation, we further analyze the quality--efficiency trade-off of Echo-DM under representative deployment-oriented configurations. Following the evaluation protocol in Section~\ref{sec:dataset-eval-protocol}, all results in this subsection are reported on the held-out \textbf{Echo-PAIR} test set. Since all variants share the same conditional latent diffusion and region-aware fusion pipeline, we focus on factors that most directly determine practical operating points, namely latent-module type, training resolution, and latent compression factor. Throughout this subsection, the number of diffusion sampling steps is fixed to 50 for fair comparison.

To better reflect realistic operating points, we report representative models from the \textbf{Echo-DM family}. Echo-DM-V is shown under three configurations, corresponding to efficiency-oriented, balanced, and quality-oriented usage scenarios. Specifically, Echo-DM-V (Efficient) uses a training resolution of \(512\times512\) with a latent compression factor of 8; Echo-DM-V (Default) uses \(512\times512\) with a compression factor of 4; and Echo-DM-V (High-Quality) uses \(1024\times1024\) with a compression factor of 4. For the two \(512\times512\) variants, we adopt high-resolution fusion at inference to compensate for detail loss introduced by low-resolution reconstruction. In contrast, the \(1024\times1024\) variant uses native-resolution fusion, since its working resolution is already close to the original image scale and therefore does not require this additional compensation. We also include \textbf{Echo-DM-R (Default)} as the framework-native RAE counterpart under the same protocol, allowing direct comparison between the VAE-based and RAE-based operating points within the same restoration-and-fusion framework.

\begin{table}[pos=htbp]
\caption{Quality--efficiency trade-off of Echo-DM on the Echo-PAIR test set under different latent-module settings, training resolutions, and latent compression factors. All variants use 50 diffusion sampling steps and share the same conditional diffusion + fusion core. The two \(512\times512\) VAE variants use high-resolution fusion at inference, while the \(1024\times1024\) VAE variant uses native-resolution fusion. Lower is better for runtime; higher is better for all quality metrics.}
\label{tab:quality_efficiency}
\centering
\scriptsize
\setlength{\tabcolsep}{3pt}
\renewcommand{\arraystretch}{1.12}
\begin{tabular*}{\tblwidth}{@{\extracolsep{\fill}} l c c c c c c c c c @{}}
\toprule
Variant & \makecell{Latent\\module} & \makecell{Training\\resolution} & \makecell{Compression\\factor} & \makecell{Fusion\\strategy} & Time$\downarrow$ & PSNR$\uparrow$ & SSIM$\uparrow$ & PSNR-ROI$\uparrow$ & SSIM-ROI$\uparrow$ \\
\midrule
Echo-DM-V (Efficient)         & VAE \citep{KingmaWelling2014VAE} & \(512\times512\)   & 8 & \makecell{High-resolution}     & \textbf{0.4324} & \textbf{40.5852} & 0.9816 & 25.0420 & 0.7982 \\
Echo-DM-V (Default)           & VAE \citep{KingmaWelling2014VAE} & \(512\times512\)   & 4 & \makecell{High-resolution}     & 1.2368 & 40.5753 & 0.9861 & 26.6898 & 0.8277 \\
Echo-DM-V (High-Quality)      & VAE \citep{KingmaWelling2014VAE} & \(1024\times1024\) & 4 & \makecell{Native-resolution} & 10.5902 & 38.4162 & \textbf{0.9870} & \textbf{26.8724} & \textbf{0.8396} \\
\midrule
Echo-DM-R       & RAE \citep{ZhengEtAl2025RAE} & \(512\times512\)   & 16 & \makecell{Native-resolution}     & 0.6180 & 31.7483 & 0.9499 & 27.6830 & 0.8218 \\
\bottomrule
\end{tabular*}
\end{table}

These rows characterize the practical quality--efficiency design space of Echo-DM under a shared restoration-and-fusion core. Within Echo-DM-V, the three settings exhibit a clear operating frontier: the Efficient configuration provides the lowest runtime, the Default configuration offers a stronger global--local balance, and the High-Quality configuration further improves structure-oriented and ROI-oriented fidelity at a substantially higher latency cost. The quality-oriented setting does not necessarily yield the highest full-image PSNR, but it achieves stronger SSIM and ROI fidelity. This pattern suggests that distortion minimization and structure-preserving restoration may favor different trade-offs.

Echo-DM-R (Default) adds a distinct RAE-based operating point to this frontier. Compared with Echo-DM-V (Default), it reduces runtime by roughly half (0.6180 vs. 1.2368 s/image) and slightly lowers ROI-SSIM, while achieving stronger ROI-PSNR than all reported VAE variants. At the same time, its full-image fidelity remains clearly below the VAE-based settings. Therefore, operating-point selection should be based on deployment priorities: Echo-DM-V is preferable when global reconstruction fidelity is most important, whereas Echo-DM-R offers a more efficient alternative when faster inference and stronger marker-region error reduction are emphasized.

\section{Discussion}
\paragraph{Practical Implications.} From a practical perspective, Echo-DM is valuable not only because it improves restoration quality, but also because it provides a deployment-friendly solution for ultrasound marker removal. In contrast to pipelines that rely on explicit intermediate masks or cascaded processing stages at inference time, Echo-DM performs restoration in an \textbf{end-to-end mask-free} manner, thereby simplifying the overall workflow and reducing dependence on additional external modules during testing. This property is particularly relevant in real-world clinical data processing, where robustness, ease of integration, and inference consistency are often as important as raw reconstruction quality. In addition, the model family can be configured at different operating points according to computational constraints. Under the same conditional-diffusion-and-fusion core, the VAE-based instantiation (Echo-DM-V) favors stronger full-image fidelity, whereas the RAE-based instantiation (Echo-DM-R) offers markedly faster inference together with stronger ROI-PSNR than the reported VAE variants. This distinction makes the framework more practically flexible under diverse deployment priorities.

\paragraph{Limitation.} The present study is conducted entirely in the ultrasound domain, and the applicability of the proposed method to other medical imaging modalities still requires further investigation. The framework nonetheless holds promise for extension to other modalities given sufficient modality-specific training data, since it is built on conditional restoration and preservation-aware fusion rather than ultrasound-specific handcrafted assumptions. A separate limitation arises when overlaid markers occupy relatively large regions or heavily overlap with subtle anatomical structures. The missing content can then be only weakly constrained by visible context, leading to inherent uncertainty in the restoration outcome. However, this limitation may be alleviated by scaling model capacity and training data, which could enable the model to learn stronger semantic and structural image priors and thereby better handle large or semantically ambiguous missing regions. This expectation is partly supported by recent progress in general-domain image inpainting, where larger-capacity models such as MAT and recent structure-aware diffusion-based approaches have shown improved completion quality for extensive missing areas and stronger semantic consistency \citep{Li_2022_CVPR,Liu_2024_CVPR}.

\paragraph{Future Work.} These limitations point to several promising directions for future research. One natural extension is to examine whether the current formulation, although developed for ultrasound marker removal, can be generalized to other medical imaging modalities with substantially different image statistics, annotation conventions, and restoration constraints. Another important direction is to relax the current supervision requirement. Since the present framework is trained on paired marked-clean data, it would be valuable to explore weaker supervision settings, such as weakly paired, semi-supervised, or partially paired learning, in order to reduce the reliance on high-quality paired annotations. Beyond its role as a benchmark for marker removal, Echo-PAIR may also support a broader range of related research problems, including marker localization, marker-aware image synthesis, and more general artifact-aware restoration in ultrasound. In this broader sense, the dataset may serve not only as a resource for marker removal, but also as a basis for studying how localized annotation artifacts can be modeled, detected, and mitigated in clinical ultrasound imaging.

\section{Conclusion}
In this work, we propose \textbf{Echo-DM}, an ultrasound marker removal framework based on conditional latent diffusion and region-aware fusion, in which a DiT-based conditional latent diffusion network is responsible for global image restoration, while the region-aware fusion module performs fidelity-oriented refinement in image space. The framework is trained in a stage-wise manner and performs end-to-end mask-free inference at test time. We further establish \textbf{Echo-PAIR}, a large-scale paired clinical ultrasound dataset that provides a representative benchmark for this task across diverse systems, views, organs, and marker types. Experimental results demonstrate that Echo-DM achieves strong restoration performance while maintaining a favorable balance between local marker removal and global background fidelity. In addition, the Echo-DM framework is compatible with both VAE-based and RAE-based latent encoder-decoder instantiations (Echo-DM-V and Echo-DM-R), which now exhibit complementary operating points in our experiments: Echo-DM-V provides stronger full-image fidelity, whereas Echo-DM-R offers substantially faster inference together with stronger marker-region PSNR. Taken together, these results suggest that ultrasound marker removal should be treated not merely as a generic image restoration problem, but as a clinically meaningful pre-processing task requiring restrained and high-fidelity editing. We hope this work can provide both a practical solution and a useful benchmark for future research on annotation artifact removal and artifact-aware restoration in medical imaging.

\appendix
\makeatletter
\@addtoreset{equation}{section}
\@addtoreset{table}{section}
\@addtoreset{figure}{section}
\makeatother
\renewcommand{\theequation}{\Alph{section}.\arabic{equation}}
\renewcommand{\thetable}{\Alph{section}.\arabic{table}}
\renewcommand{\thefigure}{\Alph{section}.\arabic{figure}}

\section{Detailed Implementation and Training Configurations}
\label{app:implementation}

\subsection{Experimental Setup}
\label{app:exp-setup}
All experiments were conducted on Ubuntu 22.04 with Python 3.10, PyTorch 2.7.1, CUDA 12.8, and cuDNN v9.10.2.21. Training and evaluation were performed on a workstation with 4 NVIDIA GeForce RTX 5090 GPUs (32 GB memory per GPU). Distributed Data Parallel (DDP) training was adopted in all stages.

Unless otherwise specified, mixed-precision training was disabled and all models were optimized in full precision. Since optimization objectives and model scales differ across stages, batch-size-related settings were configured per stage and are reported in detail in the appendix hyperparameter subsection.

\subsection{Unified Data Pipeline}
\label{app:data-pipeline}
Echo-PAIR is organized in a paired clean--marked format, where each marked image corresponds to one clean counterpart with matched file identity. Data partitioning follows a device-balanced protocol: the validation and test subsets are constructed to maintain balanced scanner-device coverage, and the remaining samples are used for training. This design reduces device-induced sampling bias in model selection and final evaluation.

Across all stages, inputs are converted to RGB and normalized to \([-1, 1]\). Bicubic interpolation is used for image resizing in the default preprocessing pipeline. Beyond this shared convention, stage-specific data handling is adopted to match the objective of each optimization stage.

In Stage I (latent-module adaptation), training does not require strict clean--marked pairing at sample level, while validation is performed on paired clean/marked data with filename intersection. In Stage II (conditional diffusion optimization), clean and marked images are strictly matched one-to-one by file stem, and invalid matches are explicitly rejected during data loading. In Stage III (region-aware fusion fine-tuning), paired clean/marked images remain the primary supervision source under the same pairing convention.

\subsection{Hyperparameter Configuration Templates}
\label{app:hyperparam-templates}
To emphasize only training-critical settings, we summarize compact model-by-parameter templates below. Non-decisive items (e.g., save paths and routine I/O arguments) are intentionally omitted.
For shorthand configuration names such as ``VAE-d4-512'' or ``d8-1024'', ``d'' denotes the latent downsample factor, and the trailing number denotes the inference resolution.

Beyond the tabulated values, we summarize stage-level resource allocation to contextualize the chosen settings. Stage~I and Stage~III are trained on a single GPU, whereas Stage~II uses DDP with 2 GPUs for 512-resolution runs and 4 GPUs for 1024-resolution runs. In practice, training cost increases markedly with spatial resolution, and 1024-resolution DiT optimization is substantially more time-consuming than its 512-resolution counterpart.

For this reason, Stage~II follows a progressive-resolution optimization schedule: lower-resolution pretraining is first performed, 512-resolution models are then warm-started, and 1024-resolution training is finally continued from the corresponding 512-resolution checkpoints. This design improves optimization stability and yields a more practical training pipeline than direct 1024-resolution training from random initialization.

\begin{table}[pos=htbp]
\caption{Stage I (latent-module adaptation) key-parameter matrix.}
\label{tab:app_stage1_key_template}
\centering
\footnotesize
\setlength{\tabcolsep}{4pt}
\renewcommand{\arraystretch}{1.12}
\begin{tabular*}{\tblwidth}{@{\extracolsep{\fill}} l c c c c c @{}}
\toprule
Parameter & VAE-d4-1024 & VAE-d8-1024 & VAE-d4-512 & VAE-d8-512 & RAE \\
\midrule
Training resolution (\texttt{image\_size/crop\_size}) & \(1024/512\) & \(1024/512\) & \(512/512\) & \(512/512\) & \(512/512\) \\
Latent downsample factor & 4 & 8 & 4 & 8 & 16 \\
Batch size (per GPU) & 2 & 2 & 4 & 4 & 2 \\
Learning rate & \(2\times10^{-5}\) & \(2\times10^{-5}\) & \(2\times10^{-5}\) & \(2\times10^{-5}\) & \(1\times10^{-4}\) \\
Max epochs & 150 & 150 & 150 & 150 & 150 \\
Early-stop patience & 30 & 30 & 30 & 30 & 30 \\
Encoder freeze strategy & Frozen encoder & Frozen encoder & Frozen encoder & Frozen encoder & Frozen encoder \\
Latent scaling factor (\texttt{scaling\_factor}) & 1.0 & 1.0 & 1.0 & 1.0 & 1.0 \\
Loss weights (\(w_{\mathrm{L1}}, w_{\mathrm{LPIPS}}, w_{\mathrm{reg}}\)) & \((1.0,0.1,10^{-6})\) & \((1.0,0.1,10^{-6})\) & \((1.0,0.1,10^{-6})\) & \((1.0,0.1,10^{-6})\) & \((1.0,0.1,10^{-6})\) \\
Mixed precision mode & fp16 & fp16 & fp16 & fp16 & bf16 \\
\bottomrule
\end{tabular*}
\end{table}
\begin{table}[pos=htbp]
\caption{Stage II (DiT optimization) key-parameter matrix.}
\label{tab:app_stage2_key_template}
\centering
\footnotesize
\setlength{\tabcolsep}{3pt}
\renewcommand{\arraystretch}{1.12}
\begin{tabular*}{\tblwidth}{@{\extracolsep{\fill}} l c c c c c @{}}
\toprule
Parameter & \makecell{Echo-DM-V\\(d4-512)} & \makecell{Echo-DM-V\\(d4-1024)} & \makecell{Echo-DM-V\\(d8-512)} & \makecell{Echo-DM-V\\(d8-1024)} & Echo-DM-R \\
\midrule
Backbone type (\texttt{model}) & DiT-S/2 & DiT-S/2 & DiT-S/2 & DiT-S/2 & \makecell{DiT with DDT head} \\
Training resolution (\texttt{image-size}) & 512 & 1024 & 512 & 1024 & 512 \\
Init. checkpoint & d4-256 ckpt\(^\ast\) & d4-512 ckpt & None & d8-512 ckpt & None \\
Batch size (per GPU) & 16 & 8 & 16 & 8 & 8 \\
Learning rate & \(1\times10^{-5}\) & \(2\times10^{-5}\) & \(1\times10^{-4}\) & \(2\times10^{-5}\) & \makecell{\(1\times10^{-4}\)\((2\times10^{-5})\)} \\
Max epochs & 300 & 300 & 300 & 300 & 800 \\
Latent scaling (\texttt{num-mul}) & 1.6 & 1.6 & 0.932 & 0.932 & / \\
Validation diffusion steps (\texttt{val-steps}) & 50 & 50 & 50 & 50 & 50 \\
Evaluation interval (\texttt{eval-epochs}) & 2 & 2 & 2 & 2 & 2 \\
Gradient checkpointing & False & True & False & True & False \\
Early-stop (metric/patience/warmup) & ROI-SSIM/50/5 & ROI-SSIM/50/5 & ROI-SSIM/50/5 & ROI-SSIM/50/5 & ROI-SSIM/50/5 \\
\bottomrule
\end{tabular*}
\end{table}
\noindent\footnotesize{\(^\ast\) The d4-256 run is a from-scratch pretraining stage at \(256\times256\), trained with DDP on 2 GPUs, per-GPU batch size \(=32\), global batch size \(=64\), learning rate \(=2\times10^{-4}\), and max epochs \(=300\). Its checkpoint is used only to initialize Echo-DM-V (d4-512) and is therefore not listed as a standalone Stage-II configuration.}\normalsize
\noindent\footnotesize{Stage II uses progressive-resolution warm-start training. Echo-DM-V (d4-512) is initialized from a d4-256 checkpoint; both 1024-resolution Echo-DM-V runs are initialized from their corresponding 512-resolution checkpoints.}\normalsize
\noindent\footnotesize{For Echo-DM-R, the learning rate is scheduled from \(1\times10^{-4}\) to \(2\times10^{-5}\) with cosine annealing.}\normalsize

\begin{table}[pos=htbp]
\caption{Stage III (fusion fine-tuning) key-parameter matrix. Each fusion model should align with one corresponding latent/diffusion configuration.}
\label{tab:app_stage3_key_template}
\centering
\footnotesize
\setlength{\tabcolsep}{4pt}
\renewcommand{\arraystretch}{1.12}
\begin{tabular*}{\tblwidth}{@{\extracolsep{\fill}} l c c c c c @{}}
\toprule  
Parameter & \makecell{Fusion\\(d4-512)} & \makecell{Fusion\\(d4-1024)} & \makecell{Fusion\\(d8-512)} & \makecell{Fusion\\(d8-1024)} & \makecell{Fusion\\(d16-512)} \\
\midrule
Paired Latent Module & VAE-d4-512 & VAE-d4-1024 & VAE-d8-512 & VAE-d8-1024 & RAE \\
Training resolution (\texttt{image-size}) & 512 & 1024 & 512 & 1024 & 512 \\
Batch size (per GPU) & 6 & 6 & 6 & 6 & 20 \\
Learning rate & \(5\times10^{-5}\) & \(5\times10^{-5}\) & \(5\times10^{-5}\) & \(5\times10^{-5}\) & \(5\times10^{-5}\) \\
Max epochs & 150 & 150 & 150 & 150 & 200 \\
Early-stop patience & 50 & 50 & 50 & 50 & 50 \\
Fusion weight (\(\alpha\)) & 0.1 & 0.1 & 0.1 & 0.1 & 0.1 \\
Noise perturbation std (\(\sigma_n\), \texttt{dit-noise-std}) & 0.05 & 0.05 & 0.05 & 0.05 & 0.05 \\
Mask dilation radius (\texttt{mask-dilate-pixels}) & 2 & 2 & 2 & 2 & 2 \\
Gradient accumulation steps & 1 & 1 & 1 & 1 & 1 \\
\bottomrule
\end{tabular*}
\end{table}

\bibliographystyle{cas-model2-names}

\bibliography{refs}

@ARTICLE{LiEtAl2024MMI,
  title={Thyroid ultrasound image database and marker mask inpainting method for research and development},
  author={Li, Xiang and Fu, Chong and Xu, Sen and Sham, Chiu-Wing},
  journal={Ultrasound in Medicine \& Biology},
  volume={50},
  number={4},
  pages={509--519},
  year={2024},
  publisher={Elsevier}
}

@ARTICLE{YingEtAl2020CMR,
  title={Cascade marker removal algorithm for thyroid ultrasound images},
  author={Ying, Xiang and Zhang, Yulin and Yu, Mei and Wei, Xi and Zhu, Jialin and Gao, Jie and Liu, Zhiqiang and Shen, Hongqian and Zhang, Ruixuan and Li, Xuewei and others},
  journal={Medical \& Biological Engineering \& Computing},
  volume={58},
  number={11},
  pages={2641--2656},
  year={2020},
  publisher={Springer}
}

@ARTICLE{YaoEtAl2020TextureSynthesis,
  title={Texture synthesis based thyroid nodule detection from medical ultrasound images: interpreting and suppressing the adversarial effect of in-place manual annotation},
  author={Yao, Siqiong and Yan, Junchi and Wu, Mingyu and Yang, Xue and Zhang, Weituo and Lu, Hui and Qian, Biyun},
  journal={Frontiers in Bioengineering and Biotechnology},
  volume={8},
  pages={599},
  year={2020},
  publisher={Frontiers Media SA}
}

@ARTICLE{SunEtAl2021EdgeCriminisi,
  title={Removal of manually induced artifacts in ultrasound images of thyroid nodules based on edge-connection and Criminisi image restoration algorithm},
  author={Sun, Ming and Meng, Qinglong and Wang, Ting and Liu, Tianci and Zhu, Ye and Qiu, Jianfeng and Lu, Weizhao},
  journal={Computer Methods and Programs in Biomedicine},
  volume={200},
  pages={105868},
  year={2021},
  publisher={Elsevier}
}

@ARTICLE{KingmaWelling2014VAE,
  title={Auto-Encoding Variational Bayes},
  author={Kingma, Diederik P and Welling, Max},
  journal={stat},
  volume={1050},
  pages={1},
  year={2014}
}

@INPROCEEDINGS{HeEtAl2022MAE,
  title={Masked autoencoders are scalable vision learners},
  author={He, Kaiming and Chen, Xinlei and Xie, Saining and Li, Yanghao and Doll{\'a}r, Piotr and Girshick, Ross},
  booktitle={Proceedings of the IEEE/CVF conference on computer vision and pattern recognition},
  pages={16000--16009},
  year={2022}
}

@ARTICLE{ZhengEtAl2025RAE,
  title={Diffusion transformers with representation autoencoders},
  author={Zheng, Boyang and Ma, Nanye and Tong, Shengbang and Xie, Saining},
  journal={arXiv preprint arXiv:2510.11690},
  year={2025}
}

@INPROCEEDINGS{ZhangEtAl2018LPIPS,
  title={The unreasonable effectiveness of deep features as a perceptual metric},
  author={Zhang, Richard and Isola, Phillip and Efros, Alexei A and Shechtman, Eli and Wang, Oliver},
  booktitle={Proceedings of the IEEE conference on computer vision and pattern recognition},
  pages={586--595},
  year={2018}
}

@ARTICLE{HoEtAl2020DDPM,
  title={Denoising diffusion probabilistic models},
  author={Ho, Jonathan and Jain, Ajay and Abbeel, Pieter},
  journal={Advances in neural information processing systems},
  volume={33},
  pages={6840--6851},
  year={2020}
}

@INPROCEEDINGS{ZhangEtAl2023ControlNet,
  title={Adding conditional control to text-to-image diffusion models},
  author={Zhang, Lvmin and Rao, Anyi and Agrawala, Maneesh},
  booktitle={Proceedings of the IEEE/CVF international conference on computer vision},
  pages={3836--3847},
  year={2023}
}

@INPROCEEDINGS{LiuEtAl2018PConv,
  title={Image inpainting for irregular holes using partial convolutions},
  author={Liu, Guilin and Reda, Fitsum A and Shih, Kevin J and Wang, Ting-Chun and Tao, Andrew and Catanzaro, Bryan},
  booktitle={Proceedings of the European conference on computer vision (ECCV)},
  pages={85--100},
  year={2018}
}

@INPROCEEDINGS{GuoEtAl2023BlindInpainting,
  title={Blind inpainting with object-aware discrimination for artificial marker removal},
  author={Guo, Xuechen and Hu, Wenhao and Ni, Chiming and Chai, Wenhao and Li, Shiyan and Wang, Gaoang},
  booktitle={ICASSP 2024-2024 IEEE International Conference on Acoustics, Speech and Signal Processing (ICASSP)},
  pages={1516--1520},
  year={2024},
  organization={IEEE}
}

@ARTICLE{ZhangEtAl2024UltrasonicNoise2Noise,
  title={Ultrasonic image's annotation removal: A self-supervised Noise2Noise approach},
  author={Zhang, Yuanheng and Jiang, Nan and Xie, Zhaoheng and Cao, Junying and Teng, Yueyang},
  journal={arXiv preprint arXiv:2307.04133},
  year={2023}
}

@INPROCEEDINGS{WangEtAl2020VCNet,
  title={Vcnet: A robust approach to blind image inpainting},
  author={Wang, Yi and Chen, Ying-Cong and Tao, Xin and Jia, Jiaya},
  booktitle={European Conference on Computer Vision},
  pages={752--768},
  year={2020},
  organization={Springer}
}

@ARTICLE{WangEtAl2004SSIM,
  title={Image quality assessment: from error visibility to structural similarity},
  author={Wang, Zhou and Bovik, Alan C and Sheikh, Hamid R and Simoncelli, Eero P},
  journal={IEEE transactions on image processing},
  volume={13},
  number={4},
  pages={600--612},
  year={2004},
  publisher={IEEE}
}

@ARTICLE{IsenseeEtAl2021nnUNet,
  title={nnU-Net: a self-configuring method for deep learning-based biomedical image segmentation},
  author={Isensee, Fabian and Jaeger, Paul F and Kohl, Simon AA and Petersen, Jens and Maier-Hein, Klaus H},
  journal={Nature methods},
  volume={18},
  number={2},
  pages={203--211},
  year={2021},
  publisher={Nature Publishing Group US New York}
}

@ARTICLE{HuynhThuGhanbari2008PSNR,
  title={Scope of validity of PSNR in image/video quality assessment},
  author={Huynh-Thu, Quan and Ghanbari, Mohammed},
  journal={Electronics letters},
  volume={44},
  number={13},
  pages={800--801},
  year={2008},
  publisher={IET}
}

@ARTICLE{WangBovik2009MSE,
  title={Mean squared error: Love it or leave it? A new look at signal fidelity measures},
  author={Wang, Zhou and Bovik, Alan C},
  journal={IEEE signal processing magazine},
  volume={26},
  number={1},
  pages={98--117},
  year={2009},
  publisher={IEEE}
}

@INPROCEEDINGS{LehtinenEtAl2018Noise2Noise,
  title={Noise2Noise: Learning Image Restoration without Clean Data},
  author={Lehtinen, Jaakko and Munkberg, Jacob and Hasselgren, Jon and Laine, Samuli and Karras, Tero and Aittala, Miika and Aila, Timo},
  booktitle={International Conference on Machine Learning},
  pages={2965--2974},
  year={2018},
  organization={PMLR}
}

@INPROCEEDINGS{LedigEtAl2017SRGAN,
  title={Photo-realistic single image super-resolution using a generative adversarial network},
  author={Ledig, Christian and Theis, Lucas and Husz{\'a}r, Ferenc and Caballero, Jose and Cunningham, Andrew and Acosta, Alejandro and Aitken, Andrew and Tejani, Alykhan and Totz, Johannes and Wang, Zehan and others},
  booktitle={Proceedings of the IEEE conference on computer vision and pattern recognition},
  pages={4681--4690},
  year={2017}
}

@INPROCEEDINGS{YuEtAl2019GatedConv,
  title={Free-form image inpainting with gated convolution},
  author={Yu, Jiahui and Lin, Zhe and Yang, Jimei and Shen, Xiaohui and Lu, Xin and Huang, Thomas S},
  booktitle={Proceedings of the IEEE/CVF international conference on computer vision},
  pages={4471--4480},
  year={2019}
}

@INPROCEEDINGS{LugmayrEtAl2022RePaint,
  title={Repaint: Inpainting using denoising diffusion probabilistic models},
  author={Lugmayr, Andreas and Danelljan, Martin and Romero, Andres and Yu, Fisher and Timofte, Radu and Van Gool, Luc},
  booktitle={Proceedings of the IEEE/CVF conference on computer vision and pattern recognition},
  pages={11461--11471},
  year={2022}
}

@ARTICLE{KawarEtAl2022DDRM,
  title={Denoising diffusion restoration models},
  author={Kawar, Bahjat and Elad, Michael and Ermon, Stefano and Song, Jiaming},
  journal={Advances in neural information processing systems},
  volume={35},
  pages={23593--23606},
  year={2022}
}

@INPROCEEDINGS{RombachEtAl2022LDM,
  title={High-resolution image synthesis with latent diffusion models},
  author={Rombach, Robin and Blattmann, Andreas and Lorenz, Dominik and Esser, Patrick and Ommer, Bj{\"o}rn},
  booktitle={Proceedings of the IEEE/CVF conference on computer vision and pattern recognition},
  pages={10684--10695},
  year={2022}
}

@ARTICLE{SahariaEtAl2021SR3,
  title={Image super-resolution via iterative refinement},
  author={Saharia, Chitwan and Ho, Jonathan and Chan, William and Salimans, Tim and Fleet, David J and Norouzi, Mohammad},
  journal={IEEE transactions on pattern analysis and machine intelligence},
  volume={45},
  number={4},
  pages={4713--4726},
  year={2022},
  publisher={IEEE}
}

@ARTICLE{HansenEtAl2024LumbarMRIInpainting,
  title={Inpainting pathology in lumbar spine MRI with latent diffusion},
  author={Hansen, Colin and Glinskis, Simas and Raju, Ashwin and Kornreich, Micha and Park, JinHyeong and Pawar, Jayashri and Herzog, Richard and Zhang, Li and Odry, Benjamin},
  journal={arXiv preprint arXiv:2406.02477},
  year={2024}
}

@INPROCEEDINGS{BenAlayaEtAl2025MedEdit,
  title={Mededit: Counterfactual diffusion-based image editing on brain mri},
  author={Alaya, Malek Ben and Lang, Daniel M and Wiestler, Benedikt and Schnabel, Julia A and Bercea, Cosmin I},
  booktitle={International Workshop on Simulation and Synthesis in Medical Imaging},
  pages={167--176},
  year={2024},
  organization={Springer}
}

@ARTICLE{ProchazkaZeman2026StableDiffusionUS,
  title={Domain adaptation of stable diffusion for ultrasound inpainting: a synthetic data approach for enhanced thyroid nodule segmentation},
  author={Prochazka, Antonin and Zeman, Jan},
  journal={Journal of Biomedical Informatics},
  pages={104963},
  year={2025},
  publisher={Elsevier}
}

@ARTICLE{YuanEtAl2025MedIENet,
  title={MedIENet: medical image enhancement network based on conditional latent diffusion model},
  author={Yuan, Weizhen and Feng, Yue and Wen, Tiancai and Luo, Guancong and Liang, Jiexin and Sun, Qianshuai and Liang, Shufen},
  journal={BMC Medical Imaging},
  volume={25},
  number={1},
  pages={372},
  year={2025},
  publisher={Springer}
}

@ARTICLE{GongEtAl2024PETDDPM,
  title={PET image denoising based on denoising diffusion probabilistic model},
  author={Gong, Kuang and Johnson, Keith and El Fakhri, Georges and Li, Quanzheng and Pan, Tinsu},
  journal={European Journal of Nuclear Medicine and Molecular Imaging},
  volume={51},
  number={2},
  pages={358--368},
  year={2024},
  publisher={Springer}
}

@INPROCEEDINGS{PeeblesXie2023DiT,
  title={Scalable diffusion models with transformers},
  author={Peebles, William and Xie, Saining},
  booktitle={Proceedings of the IEEE/CVF international conference on computer vision},
  pages={4195--4205},
  year={2023}
}

@article{xu2025diffcnn,
  title={DiffCNN: A collaborative framework of diffusion model and CNN for semi-supervised medical image segmentation},
  author={Xu, Shanshan and Tian, Lixia},
  journal={Neural Networks},
  volume={191},
  pages={107813},
  year={2025},
  publisher={Elsevier}
}

@article{wang2024narrowing,
  title={Narrowing the semantic gaps in u-net with learnable skip connections: The case of medical image segmentation},
  author={Wang, Haonan and Cao, Peng and Yang, Jinzhu and Zaiane, Osmar},
  journal={Neural Networks},
  volume={178},
  pages={106546},
  year={2024},
  publisher={Elsevier}
}

@InProceedings{Li_2022_CVPR,
  title={Mat: Mask-aware transformer for large hole image inpainting},
  author={Li, Wenbo and Lin, Zhe and Zhou, Kun and Qi, Lu and Wang, Yi and Jia, Jiaya},
  booktitle={Proceedings of the IEEE/CVF conference on computer vision and pattern recognition},
  pages={10758--10768},
  year={2022}
}

@InProceedings{Liu_2024_CVPR,
  title={Structure matters: Tackling the semantic discrepancy in diffusion models for image inpainting},
  author={Liu, Haipeng and Wang, Yang and Qian, Biao and Wang, Meng and Rui, Yong},
  booktitle={Proceedings of the IEEE/CVF Conference on Computer Vision and Pattern Recognition},
  pages={8038--8047},
  year={2024}
}

@article{meng2025dfcl,
  title={DFCL: Dual-pathway fusion contrastive learning for blind single-image visible watermark removal},
  author={Meng, Bin and Zhou, Jiliu and Yang, Haoran and Liu, Jiayong and Pu, Yifei},
  journal={Neural Networks},
  volume={184},
  pages={107077},
  year={2025},
  publisher={Elsevier}
}

\end{document}